\pdfoutput=1

\documentclass[11pt]{article}

\usepackage[]{EMNLP2023}

\usepackage{times}
\usepackage{latexsym}

\usepackage[T1]{fontenc}

\usepackage[utf8]{inputenc}

\usepackage{microtype}
\usepackage{mathtools}
\usepackage{tikz}
\usepackage{multirow}
\usepackage{booktabs}
\usepackage{bbm}
\usepackage{graphicx}
\usepackage{amsthm}
\usepackage{tikz}
\usepackage{enumitem}

\theoremstyle{definition}
\newtheorem{definition}{Definition}[section]
\usetikzlibrary{shapes,decorations,arrows,calc,arrows.meta,fit,positioning}
\tikzset{
    -Latex,auto,node distance =1 cm and 1 cm,semithick,
    state/.style ={ellipse, draw, minimum width = 0.7 cm},
    point/.style = {circle, draw, inner sep=0.04cm,fill,node contents={}},
    bidirected/.style={Latex-Latex,dashed},
    el/.style = {inner sep=2pt, align=left, sloped}
}

\title{Counterfactual Augmentation for Multimodal Learning Under Presentation Bias}

\author{Victoria Lin\footnotemark[1] \\
  Carnegie Mellon University \\
  \texttt{vlin2@andrew.cmu.edu} \\\And
  Louis-Philippe Morency \\
  Carnegie Mellon University \\
  \texttt{morency@cs.cmu.edu} \\\AND
  Dimitrios Dimitriadis \\
  Microsoft Research \\
  \texttt{didimit@microsoft.com} \\\And
  Srinagesh Sharma \\
  Microsoft \\
  \texttt{srsharm@microsoft.com}
}

\begin{document}
\maketitle
\begin{abstract}
In real-world machine learning systems, labels are often derived from user behaviors that the system wishes to encourage. Over time, new models must be trained as new training examples and features become available. However, feedback loops between users and models can bias future user behavior, inducing a \textit{presentation bias} in the labels that compromises the ability to train new models. In this paper, we propose \textit{counterfactual augmentation}, a novel causal method for correcting presentation bias using generated counterfactual labels. Our empirical evaluations demonstrate that counterfactual augmentation yields better downstream performance compared to both uncorrected models and existing bias-correction methods. Model analyses further indicate that the generated counterfactuals align closely with true counterfactuals in an oracle setting.
\end{abstract}

\section{Introduction}
\label{sec:intro}\

\renewcommand{\thefootnote}{\fnsymbol{footnote}}
\footnotetext[1]{The majority of this work was conducted while this author was an intern at Microsoft.}
\renewcommand{\thefootnote}{\arabic{footnote}}

\begin{figure}[!ht]
    \centering
    \includegraphics[width=\columnwidth]{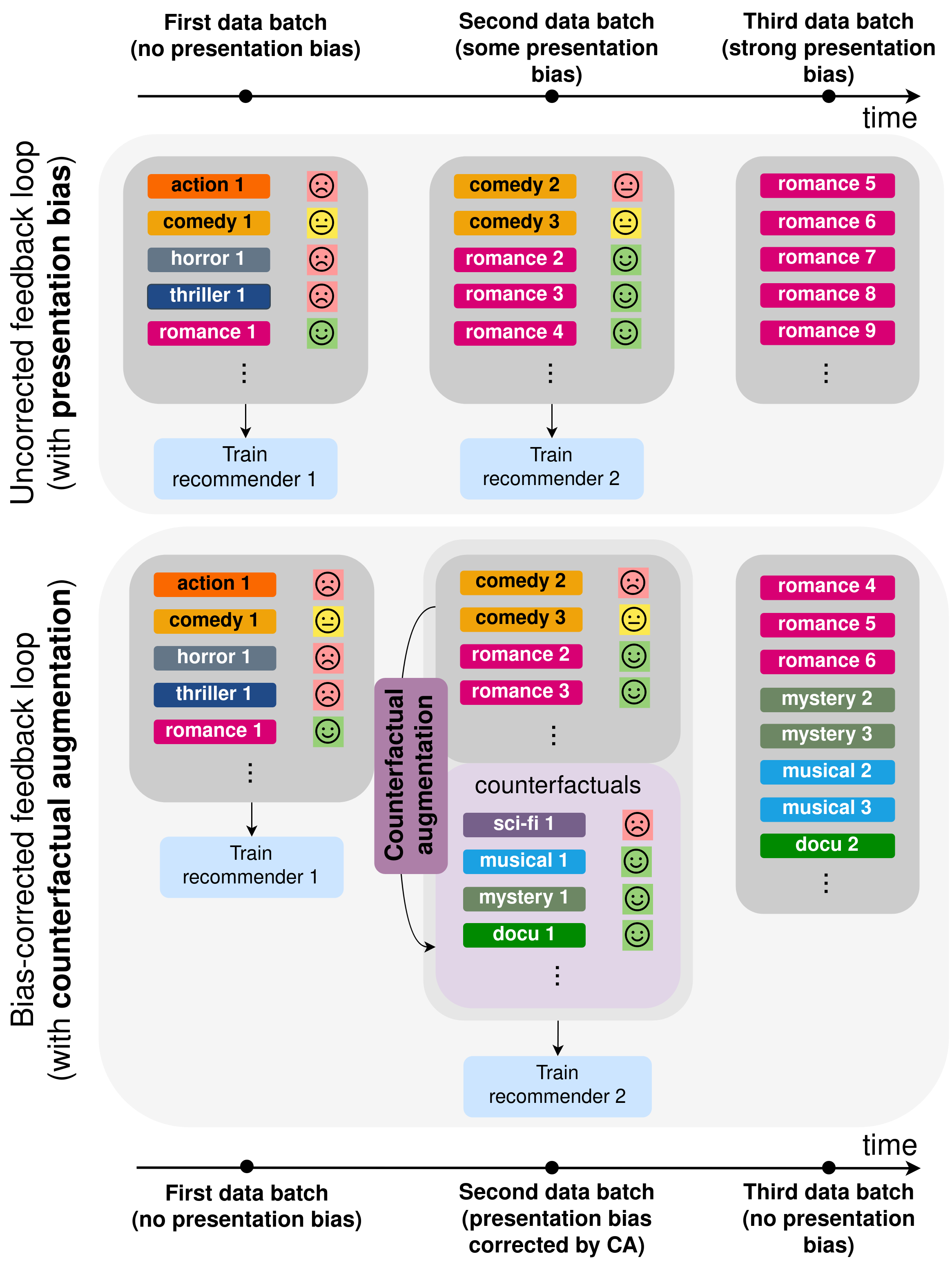}
    \caption{An illustration of how presentation bias may arise from feedback loops (e.g., in a movie recommendation system). The top sequence depicts uncorrected presentation bias, while the bottom sequence demonstrates how our method, \textit{counterfactual augmentation}, can correct presentation bias.}
    \label{fig:counterfactual_augmentation_concept}
\end{figure}

Deployment of machine learning models is ubiquitous in the real world, ranging from web search ranking to movie recommendation. To ensure good performance, new models must be trained periodically, since new training examples may become available, and the types of features that are collected can evolve over time (e.g., from tabular to multimodal data). 
For user-facing models like recommenders, labels are often derived from user behaviors that the model wishes to encourage, and user-model interactions continuously produce new data that can be used for training models. Modern NLP, for instance, relies heavily on models that learn from user feedback, not the least of which are ChatGPT and other large language models that comprise the current state-of-the-art.

In practice, however, feedback loops between the user and the model can influence future user behavior, inducing \textit{presentation bias} over the labels \citep{joachims2017unbiased, pan2021correcting}. By shifting the label distribution away from users' true preferences, presentation bias compromises the ability to train new models \citep{schmit2018human, krauth2020offline}. For example, an algorithm may present future content based on a user's interactions with prior content. As the user engages with the algorithm's recommendations or outputs, it will recommend more of that same type of content---even if there are other types of content the user might also enjoy (Figure \ref{fig:counterfactual_augmentation_concept}). 


Presentation bias negatively affects the data distribution in two major ways. (1) \textit{Bias amplification.} New labels are dependent on the prior behavior of the model, so they may not reflect the user's true preferences. This bias will amplify as more training loops are completed on biased data. (2) \textit{Label homogenization.} As the model learns user behaviors, most users' responses to its recommendations will be positive, so variation in user feedback decreases. 


In this paper, we aim to correct the presentation bias resulting from feedback loops. We first propose that presentation bias arises due to the causal relationship between a model's recommendations and a user's behavior, which affects \textit{which labels are observed}. Users tend to interact with recommended items, so under presentation bias, we are more likely to observe labels for recommended items---while without presentation bias, users would interact with all items (or a random subset), so labels would be observed for the full distribution. We conclude that we can break the causal link behind presentation bias with a \textit{counterfactual} question: how would users have reacted \textit{had they interacted with all items}, contrary to reality?


With this idea as our foundation, we introduce \textit{counterfactual augmentation} (Figure \ref{fig:counterfactual_augmentation_concept}), a causal approach for reducing presentation bias using generated counterfactual labels.
\footnote{Our code is publicly available at \url{https://github.com/microsoft/causaltransfer}.}
Because ``true'' counterfactuals are by definition unknown,
counterfactual augmentation leverages the causal relationship between the model's behavior and the user's behavior to generate realistic counterfactual labels. We generate counterfactuals for the labels that are unobserved due to presentation bias, then augment the observed labels with the generated ones. Intuitively, this supplies labels over the full data distribution, yielding a bias-corrected dataset.

We evaluate our method on predictive tasks in language and multimodal settings that reflect real-world presentation bias. We consider data with evolving feature spaces, where over time the features transition from simpler features to richer language or multimodal ones.\footnote{Although we choose this setting to more accurately represent real-world data, counterfactual augmentation does not require an evolving feature space. In the appendix (Section \ref{sec:tabonly_results}), we empirically show the effectiveness of counterfactual augmentation in data settings without feature evolution.} In our experiments, we demonstrate that counterfactual augmentation effectively corrects presentation bias when training predictive models, outperforming both uncorrected models and existing bias correction methods. We conduct model analyses that examine why counterfactual augmentation is effective for reducing presentation bias and discover that our generated counterfactuals align closely with true counterfactuals in an oracle setting.

\section{Problem Statement}
\label{sec:problem}

\begin{figure}
    \centering
     \scalebox{1.0}[1.0]{\begin{tikzpicture}
        \node[state] (xtab1) at (0,0) {$X_0$};
        \node[state] (y1) [left =of xtab1] {$Y_0$};
    
        \node[state] (r1) [below =of y1] {$R_0$};
        \node[state, red] (a2) [right =of r1] {$A$};
        \node[state] (y2) [right =of xtab1] {$Y$};
        \node[state] (xtab2) [right =of y2] {$X$};
        \node[state, fill=white!70!green] (r2) [right =of a2] {$R$};
        \node[state] (xmultimodal) [right =of r2] {$W$};
    
        \path (xtab1) edge (r1);
        \path (y1) edge (r1);
        \path[red] (r1) edge (a2);
        \path[red] (a2) edge (y2);
        \path (xtab2) edge (r2);
        \path (xmultimodal) edge (r2);
        \path (y2) edge (r2);
    
    \end{tikzpicture}}
    \caption{Proposed mechanism of presentation bias. $X_t$, $W_t$, and $Y_t$ denote simple features, rich features, and labels at time $t$ (no subscript for $t=1$), while $R_t$ is a model (e.g., a recommender) trained over the input features and labels. $A$ indicates which items a user interacts with.}
    \label{fig:causal_mechanism}
\end{figure}
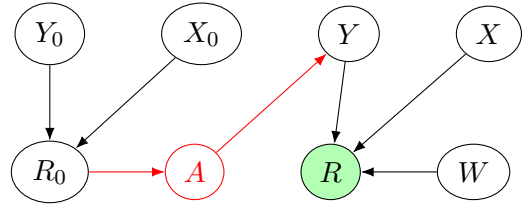


We formalize the problem of presentation bias in machine learning systems in causal terms (Figure \ref{fig:causal_mechanism}). These systems usually consume both simple features, such as metadata, and \textit{rich} features, such as text or images, training on user interactions with different items to produce recommendations. 

Let $t$ be a time index, and let \(X_t\) denote the simple features defined over the feature space \(\mathbf{X}\). Similarly, let \(W_t\) denote rich features defined over the feature space \(\mathbf{W}\). We denote \textit{true} user item preferences as \(Y_t \in \mathbf{Y}\) and \textit{predicted} user item preferences as \(R\) (for simplicity, we can think of these as binary recommendations). Finally, let $A_t$ be an indicator of which items the user interacts with. Due to feature evolution, only $X_t$ is observed at earlier time points, while later both $X_t$ and $W_t$ are observed. Without loss of generality, assume that the feature evolution occurs in the first two time points $t=0$ and $t=1$. For ease of notation, we do not include time index subscripts when $t=1$.

At $t=0$, a predictive model $R_0$ is trained on an observed feature set $X_0$ and labels $Y_0$. This model makes predictions about user preferences for unseen items and recommends items to the user. These recommendations influence $A$---which of those items the user subsequently interacts with---because users are much more likely to interact with recommended items, such that $P(A=1 | R_0=0) \ll P(A=1|R_0=1)$. In turn, this induces a presentation bias in the distribution of observed $Y$, the user's measured preferences at $t=1$. Due to the presentation bias, there is a very high probability of observing $Y$ when $R_0=1$ and a very low probability of observing $Y$ when $R_0=0$. 

At $t=1$, a full set of simple and rich features ($X$, $W$) is observed due to feature evolution. However, because the distribution of $Y$ has been influenced by $R_0$, a second model $R$ trained on $X$, $W$, and $Y$ will not correctly learn user preferences.

\textbf{Example.} Consider a system that must categorize a user's emails as important and unimportant. $X$ is email metadata, and $W$ is the text of the email. $Y$ is an indicator of whether the user interacted with the email positively (e.g., replied) or negatively (e.g., reported spam). $R_0$ is a classifier trained on $X_0$ and $Y_0$ to label emails as important or unimportant. Users preferentially interact with important emails, so emails with $R_0=1$ have a much higher chance of user interaction (i.e., $A=1$) and therefore of having an observed label $Y$, inducing a bias in $Y$ that depends on $R_0$. After $R_0$ is trained, the system's administrators want to train a new, improved model $R$ using both $X$ and $W$. However, the bias in $Y$ will affect the ability to train $R$.

\section{Methods}
\label{sec:methods}

To eliminate presentation bias, we notice that we must block the causal path between $R_0$ and $Y$ so that $R_0$ no longer influences which $Y$ are observed. Because these two variables are linked by the \textit{mediator} $A$, we can block the path by controlling for $A$. To do so, we define the \textit{counterfactual} $Y^{A=a}$, the value $Y$ would have taken \textit{had} $A=a$. 

    

\subsection{Counterfactual augmentation}

Using $Y^{A=a}$, we block the path between the recommender $R_0$ and the label $Y$ with the following intuition. $A$ indicates which items users interact with and thus which labels are observed. We can therefore eliminate the influence of $A$ by generating a synthetic data distribution in which \textit{all} items receive user interaction and \textit{all} $Y$ are ``observed.''



Formally, let $P(Y)$ denote the marginal distribution of the labels and $P(X,Y)$ denote the joint marginal distribution of the features and the labels. In an unbiased setting, a model $f$ is optimized over data $(x,y) \sim P(X,Y)$. Under presentation bias, however, only a portion of $P(Y)$ is observed: the conditional distribution $P(Y|A=1)$. Consequently, the model $f$ is trained over data $(x,y) \sim P(X,Y|A=1)$, which may lead to convergence to a non-optimal solution.

\begin{definition}[Counterfactual augmentation]
To correct presentation bias in the data distribution, counterfactual augmentation creates an approximation of the marginal label distribution $P(Y)$ using the estimated distribution of \textit{counterfactual} labels $Y^{A=1}$, or what $Y$ would have been had $A=1$. This allows us to define $P_{CA}(Y)$, a counterfactually augmented marginal label distribution:
\begin{align*}
    \begin{split}
    P_{CA}(Y)&=P(Y|A=1)P(A=1)\\
    &+\underbrace{\hat{P}(Y^{A=1}|A=0)P(A=0)}_{\text{counterfactual augmentation}}
    \end{split}
\end{align*}

Combining labels from $P_{CA}(Y)$ with the known features, we have $P_{CA}(X,Y)$, a counterfactually augmented marginal data distribution:
\begin{align*}
    \begin{split}
    P_{CA}(X,Y)&=P(X,Y|A=1)P(A=1)\\
    &+\underbrace{P(X,\hat{Y}^{A=1}|A=0)P(A=0)}_{\text{counterfactual augmentation}}
    \end{split}
\end{align*}
\end{definition}

From $P_{CA}(X,Y)$, bias-corrected data can be sampled, such that the model $f$ is now optimized over $(x,y)\sim P_{CA}(X,Y)$. Supposing $P_{CA}(X,Y)$ is a good approximation of $P(X,Y)$, $f$ should converge to a near-optimal solution.

\subsection{Multimodal counterfactual GAN}

\begin{figure*}[!ht]
    \centering
    \includegraphics[width=\textwidth]{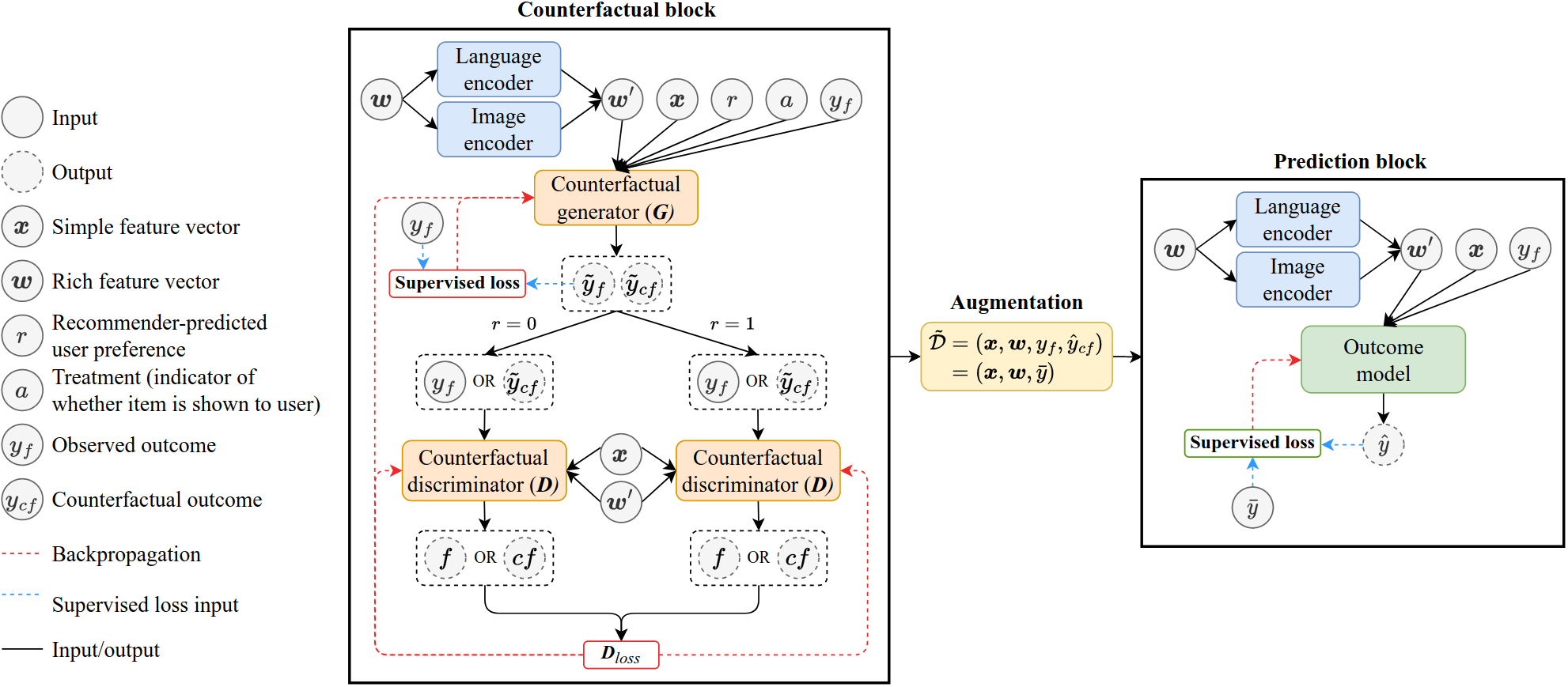}
    \caption{Diagram of our multimodal counterfactual GAN architecture. In the \textit{counterfactual block}, a generator $G$ takes multimodal data as input and generates a factual label $\tilde{y}_f$ and a counterfactual label $\tilde{y}_{cf}$. As the true factual label $y_f$ is known, it is used to learn a supervised loss between $y_f$ and $\tilde{y}_f$ that helps to train $G$. At random, either the true factual label $y_f$ or the generated counterfactual label $\tilde{y}_{cf}$ is passed to a discriminator $D$, conditional on the recommendation $r$ corresponding to the label. The discriminator must determine whether the label it has received is factual or counterfactual, and its loss $D_{loss}$ is used to further train both $G$ and $D$. After the GAN has been trained, its counterfactuals are used to augment data that is used for predictive tasks (e.g., \textit{prediction block}).}
    \label{fig:counterfactual_gan}
\end{figure*}


We implement counterfactual augmentation with a generative adversarial network (GAN) capable of generating realistic counterfactual labels given multimodal input data.\footnote{We note that the main novelty of our paper is the principle of counterfactual augmentation for correcting presentation bias, rather than the particular means by which the counterfactual is estimated. Empirically, however, we found our counterfactual GAN implementation to generalize well. 
} Inspired by the work of \citet{yoon2018ganite}, who propose a GAN (GANITE) specifically for estimating individual causal effects, we generate labels---both factual and counterfactual---with a generator $G$, then train a discriminator $D$ to distinguish factual from counterfactual labels. Our architecture (Figure \ref{fig:counterfactual_gan}) extends their work in several core aspects:


\textbf{Mediators.} Rather than estimating the direct effect of an intervention $A$ on an outcome $Y$, we seek to model the indirect effect of a variable $R$ on an outcome $Y$ through the \textit{mediator} $A$. We account for both of these dependencies, allowing us to later block the effect of $R$ on $Y$ by intervening on $A$. 

\textbf{Language and multimodal data.} Where GANITE was designed for tabular data only, our implementation handles richer features like text and images. We integrate language and image encoders into the architecture that can be simultaneously fine-tuned as the counterfactual GAN is trained.

\textbf{Correcting the discriminator constraint.} Let $Y_{cf}$ denote a counterfactual label and $Y_f$ denote a factual label. The discriminator of GANITE encourages 
$\hat{P}(Y_{cf}|X)\rightarrow P(Y_f|X)$, 
i.e., the estimated distribution of counterfactual labels should converge to the true distribution of factual labels. 
However, in feedback loops, the label is much more likely to be observed (i.e., factual) when $R=1$ than when $R=0$. Then the discriminator may enforce $\hat{P}(Y_{cf}|X,R=0) \rightarrow P(Y_f|X,R=1)$, which would mean that labels follow the same distribution regardless of whether $R=1$ or $R=0$. 

We address this problem by defining two separate discriminators---one for each recommendation condition. Each discriminator is arbitrarily passed a factual or counterfactual label from its  recommendation condition, and it must identify whether the label is factual or counterfactual. The separate discriminators encourage the realistic constraints $\hat{P}(Y_{cf}|X,R=0)\rightarrow P(Y_f|X,R=0)$ and $\hat{P}(Y_{cf}|X,R=1)\rightarrow P(Y_f|X,R=1)$.


\section{Experiments}
\label{sec:experiments}

We conduct empirical evaluations to assess how well counterfactual augmentation corrects presentation bias, with the aim of improving downstream performance. We evaluate on predictive machine learning tasks, which reflect real-world models' goals of predicting user behavior, and in multiple data settings, both synthetic and real-world. To facilitate detailed analysis of our models, we introduce a procedure for inducing realistic presentation bias in unbiased datasets. All data and code for our experiments will be publicly released.

\subsection{Datasets}


To recreate feature evolution in our experiments, we evaluate on datasets that contain both tabular features and rich features like text or images. We select two datasets from the Multimodal AutoML Benchmark \citep{shi2021benchmarking}: Airbnb and Clothing. \textbf{Airbnb} consists of 22,895 Airbnb listings for the city of Melbourne, including metadata, text descriptions, and images of the property. The nightly price of the listing is the label. \textbf{Clothing} comprises 23,486 reviews of women's clothing from an online retailer, with metadata, the title, and the text of the review. The review score is treated as the label. Both labels can be predicted directly via regression, but they can also be discretized to be used in classification tasks (as we do in our evaluation). For our binary classification tasks, we binarize both datasets in a 0-1 proportion of approximately 0.25 to 0.75 to reflect real-world data, in which the majority of feedback received from users is positive.


We further create a synthetic version of the Airbnb dataset (\textbf{Synthetic}) in which the features are taken from the real dataset, but the label is synthesized as a noisy function of the tabular features and the multimodal features. The purpose of this dataset is to evaluate the efficacy of counterfactual augmentation in a ``best-case scenario'' in which we know that there is some signal about the label that can be gained independently from both the simple features and the rich features. We use a binary label, again to reflect a ``best-case scenario'' in which the downstream task is relatively easy (compared to multi-class classification or regression), with a 0-1 proportion of approximately 0.25 to 0.75.

Additional details about the datasets are provided in the appendix (Section \ref{sec:data_details}).

\subsection{Method for inducing presentation bias}

To induce presentation bias in these datasets in a way that will allow for post-hoc model analysis, we use a procedure that mimics feedback loops in real-world systems. We first create three splits of the data, which correspond to the three data batches in Figure \ref{fig:counterfactual_augmentation_concept}. We refer to these splits as $D_{original}$, $D_{train}$, and $D_{eval}$. On $D_{original}$, which has no presentation bias, we fit a model $M_{tab}$ on the labels $Y_{original}$, using only tabular features.

We use $M_{tab}$ to predict labels $R_{train}$ for $D_{train}$ using tabular features, where $R_{train} = M_{tab}(X_{train})$. $R_{train}$ corresponds to $R_0$ in our causal structure. Next, we drop 90\% of the labels from samples in $D_{train}$ where $R_{train}=0$ (where $Y$ is multi-class or binary, we use a threshold value instead). This induces presentation bias by creating the causal dependency $R_0 \rightarrow A \rightarrow Y$, where labels are observed with high probability when $R_0=1$ and with low probability when $R_0=0$. We also randomly drop $\sim$35\% of samples from $D_{train}$ with equal probability (reflecting the remaining items that users do not interact with).

Finally, for $D_{eval}$, we create an \textbf{unbiased} version in which we leave $D_{eval}$ as it is, and a \textbf{biased} version $D_{biased}$. For $D_{biased}$, we again use $M_{tab}$ to predict labels $R_{eval}$ using only tabular features, where $R_{eval} = M_{tab}(X_{eval})$. We then drop 90\% of the samples in $D_{biased}$ where $R_{eval}=0$.

\subsection{Models}

\textbf{Baselines.} We compare counterfactual augmentation against several baselines. First, we include a model without bias correction (\textbf{Uncorrected}). To provide the best chance of achieving good performance, we use pre-trained transformer architectures fine-tuned on the respective task datasets: DistilBERT \citep{sanh2020distilbert} for language and ViT \citep{dosovitskiy2021image} for images. These models are used as encoders for the text and images of the datasets. Once embeddings are obtained, they are concatenated with the tabular data and passed to a final layer fine-tuned for the predictive task.


Our remaining baselines are implementations of existing methods for correcting presentation bias, both of which we describe further in Section \ref{sec:related_work}. In our experiments, the \textbf{IPW} baseline is implemented identically to the uncorrected baseline; however, when fine-tuning the final task layer, an \textit{inverse propensity weighted} loss \citep{wang2016learning} is used.
The \textbf{Dragonnet} baseline is an adaptation of a method proposed by \citet{shi2019adapting} for jointly estimating causal treatments and outcomes with a single neural network. To make this method compatible with our data setting, we pre-embed the text and images before passing them to Dragonnet, and we also modify the final layer to output estimated counterfactuals rather than estimated causal effects.

\textbf{Counterfactual augmentation.} In our proposed method, counterfactual augmentation (\textbf{CA}), we train our multimodal counterfactual GAN on a biased dataset, then use the GAN to generate the counterfactual labels for all samples for which labels are not observed. Combining the generated labels with the observed labels, we have a bias-corrected dataset. With this bias-corrected data, we encode text and images using fine-tuned DistilBERT and ViT, combine them with the tabular data, and train a final layer for the specific task.

Additional details about the training procedures are provided in the appendix (Section \ref{sec:training_details}).

\begin{table*}[!ht]
    \centering
    \scalebox{0.73}{
    \begin{tabular}{c|cccc|cccc|cccc}
    \toprule
    \toprule
        & \multicolumn{4}{c|}{Synthetic} & \multicolumn{4}{c|}{Airbnb} & \multicolumn{4}{c}{Clothing} \\
        & Acc. & $F_1$ & ${F_1}_{mac}$ & ${F_1}_{min}$ & Acc. & $F_1$ & ${F_1}_{mac}$ & ${F_1}_{min}$ & Acc. & $F_1$ & ${F_1}_{mac}$ & ${F_1}_{min}$ \\
        \midrule
        Uncorrected & 81.7 & 76.9 & 58.4 & 27.2 & 84.8 & 82.5 & 69.5 & 48.0 & 77.5 & 68.1 & 45.5 & 3.8 \\
        IPW & 82.0 & 78.4 & 62.0 & 34.4 & 86.0 & 84.7 & 74.2 & 56.8 & 79.6 & 73.8 & 56.8 & 25.4 \\
        Dragonnet & 81.9 & 79.8 & 65.9 & 42.5 & 81.8 & 79.9 & 65.7 & 42.1 & 77.1 & 67.1 & 43.5 & 0.0 \\
        CA (ours) & \textbf{82.7} & \textbf{82.0} & \textbf{71.1} & \textbf{52.8} & \textbf{87.4} & \textbf{87.4} & \textbf{80.2} & \textbf{68.2} & \textbf{80.3} & \textbf{76.7} & \textbf{63.1} & \textbf{37.9} \\
        \midrule
        Improvement & 0.7\% & 2.2\% & 5.2\% & 10.3\% & 1.5\% & 2.7\% & 6.0\% & 11.5\% & 0.7\% & 2.9\% & 6.3\% & 12.5\% \\
        \midrule
        Oracle & 82.6 & 79.7 & 64.8 & 39.7 & 88.5 & 88.4 & 81.8 & 70.7 & 80.1 & 74.6 & 58.3 & 28.2 \\
    \bottomrule
    \bottomrule
    \end{tabular}
    }
    \caption{Results on binary classification tasks (unbiased evaluation dataset).}
    \label{tab:results_classification_unbiased}
\end{table*}

\begin{table*}[!ht]
    \centering
    \scalebox{0.73}{
    \begin{tabular}{c|cccc|cccc|cccc}
    \toprule
    \toprule
        & \multicolumn{4}{c|}{Synthetic} & \multicolumn{4}{c|}{Airbnb} & \multicolumn{4}{c}{Clothing} \\
        & Acc. & $F_1$ & ${F_1}_{mac}$ & ${F_1}_{min}$ & Acc. & $F_1$ & ${F_1}_{mac}$ & ${F_1}_{min}$ & Acc. & $F_1$ & ${F_1}_{mac}$ & ${F_1}_{min}$ \\
        \midrule
        Uncorrected & \textbf{85.1} & 80.5 & 54.9 & 17.9 & 88.2 & 86.1 & 67.1 & 40.8 & 79.9 & 71.3 & 45.7 & 2.5 \\
        IPW & 85.0 & 81.0 & 56.9 & 22.0 & 88.5 & 87.1 & 70.7 & 47.8 & \textbf{81.4} & \textbf{75.7} & \textbf{55.5} & \textbf{21.5} \\
        Dragonnet & 81.0 & 79.3 & 57.1 & 25.0 & 86.4 & 81.1 & 51.4 & 10.2 & 79.6 & 70.6 & 44.3 & 0.0 \\
        CA (ours) & 84.9 & \textbf{83.0} & \textbf{64.0} & \textbf{36.6} & \textbf{88.7} & \textbf{88.2} & \textbf{74.5} & \textbf{55.6} & 80.5 & 74.6 & 53.6 & 18.2 \\
    \midrule
        Improvement & - & 2.0\% & 6.9\% & 11.6\% & 0.2\% & 1.1\% & 3.9\% & 7.8\% & - & - & - & - \\
        \midrule
        Oracle & 85.1 & 81.6 & 58.8 & 25.9 & 89.7 & 89.5 & 77.9 & 61.7 & 81.8 & 76.3 & 56.8 & 24.0 \\
    \bottomrule
    \bottomrule
    \end{tabular}
    }
    \caption{Results on binary classification tasks (biased evaluation dataset).}
    \label{tab:results_classification_biased}
\end{table*}

\begin{table}[!ht]
    \centering
    \resizebox{\columnwidth}{!}{
    \begin{tabular}{c|cc|cc|ccc}
    \toprule
    \toprule
        & \multicolumn{4}{c|}{Regression} & \multicolumn{3}{c}{Multi-class} \\
        \cmidrule{2-8}
        & \multicolumn{2}{c|}{Airbnb} & \multicolumn{2}{c|}{Clothing} & \multicolumn{3}{c}{Clothing} \\
        & $R^2$ & RMSE & $R^2$ & RMSE & Acc. & $F_1$ & ${F_1}_{mac}$ \\
        \midrule
        Uncorrected & 0.126 & 0.935 & 0.092 & 0.953 & 55.7 & 39.9 & 14.4 \\
        IPW & 0.127 & 0.934 & 0.120 & 0.938 & 56.0 & 40.9 & 15.7 \\
        Dragonnet & - & - & - & - & - & - & - \\
        CA (ours) & \textbf{0.186} & \textbf{0.902} & \textbf{0.197} & \textbf{0.896} & \textbf{57.1} & \textbf{44.6} & \textbf{20.1} \\
        \midrule
        Improvement & 5.9\% & 3.2\% & 7.6\% & 4.1\% & 1.1\% & 3.7\% & 4.4\% \\
        \midrule
        Oracle & 0.131 & 0.932 & 0.194 & 0.898 & 57.2 & 44.3 & 19.2 \\
    \bottomrule
    \bottomrule
    \end{tabular}
    }
    \caption{Results on regression and multi-class classification tasks (unbiased evaluation dataset). Our RMSE metric is normalized RMSE, or RMSE divided by the standard deviation of the evaluation set.}
    \label{tab:results_regression_multiclass_unbiased}
\end{table}

\begin{table}[!ht]
    \centering
    \resizebox{\columnwidth}{!}{
    \begin{tabular}{c|cc|cc|ccc}
    \toprule
    \toprule
        & \multicolumn{4}{c|}{Regression} & \multicolumn{3}{c}{Multi-class} \\
        \cmidrule{2-8}
        & \multicolumn{2}{c|}{Airbnb} & \multicolumn{2}{c|}{Clothing} & \multicolumn{3}{c}{Clothing} \\
        & $R^2$ & RMSE & $R^2$ & RMSE & Acc. & $F_1$ & ${F_1}_{mac}$ \\
        \midrule
        Uncorrected & 0.112 & 0.942 & 0.083 & 0.958 & 58.4 & 43.2 & 14.9 \\
        IPW & 0.112 & 0.942 & 0.098 & 0.950 & \textbf{58.6} & 43.8 & 15.8 \\
        Dragonnet & - & - & - & - & - & - & - \\
        CA (ours) & \textbf{0.134} & \textbf{0.930} & \textbf{0.128} & \textbf{0.934} & 58.6 & \textbf{43.9} & \textbf{16.1} \\
        \midrule
        Improvement & 2.2\% & 1.2\% & 3.0\% & 1.6\% & - & 0.1\% & 0.3\% \\
        \midrule 
        Oracle & 0.108 & 0.944 & 0.156 & 0.919 & 59.3 & 46.0 & 18.2 \\
    \bottomrule
    \bottomrule
    \end{tabular}
    }
    \caption{Results on regression and multi-class classification tasks (biased evaluation dataset).}
    \label{tab:results_regression_multiclass_biased}
\end{table}

\subsection{Evaluation}

In our evaluation, our models are fit on $D_{train}$ (which contains presentation bias) and evaluated on both $D_{eval}$ and $D_{biased}$.
Evaluation on $D_{eval}$ indicates how well our model predicts the label in a setting where presentation bias is not a factor (i.e., if we knew all labels). Evaluation on $D_{biased}$ indicates how well our model predicts the label given the data that is available to us in reality.

We note that as a consequence of presentation bias, any class or distribution imbalance in the labels $Y$ will be amplified, since there is a positive relationship between the predicted labels $R$ and the true labels $Y$. 
This imbalance reflects the real-world tendency for users to like their recommendations and for positive labels to dominate. Therefore, in classification tasks, overall accuracy and $F_1$ score will be artificially high for a model that simply predicts the most common class. Important measures of success for a method will instead be $F_{1_{mac}}$, or macro $F_1$ score ($F_1$ score uniformly weighted across all classes), and for binary classification, $F_{1_{min}}$, or $F_1$ score on the minority class.

\section{Results and Discussion}
\label{sec:results}

\subsection{Prediction task results}

We evaluate counterfactual augmentation against our baselines in the Synthetic, Airbnb, and Clothing data settings on binary classification tasks (Tables \ref{tab:results_classification_unbiased} and \ref{tab:results_classification_biased}) and multi-class classification and regression tasks (Tables \ref{tab:results_regression_multiclass_unbiased} and \ref{tab:results_regression_multiclass_biased}). ``Improvement'' is computed by taking the difference between the CA score and the score of the next-best method for that metric, which is generally IPW.

We observe that when evaluating in an unbiased setting (which reflects ``true'' preferences), counterfactual augmentation offers the best performance across all metrics for all tasks on all datasets, often by a significant margin. It outperforms not only the uncorrected baseline but also both bias-correction baselines, IPW and Dragonnet. When evaluating in a biased setting (which reflects the evaluation data we have available in reality), counterfactual augmentation also improves performance across metrics, tasks, and datasets. In general, it outperforms competing bias-correction methods, with the single exception being the binary classification task for the Clothing Review dataset, where it does not achieve as much improvement as IPW but still offers substantial gains over the uncorrected model.

Importantly, the biggest improvements resulting from counterfactual augmentation are in the minority classes. As we mention previously, due to the imbalance in the distribution of $Y$, macro and minority class $F_1$ score are the best measures of performance. Furthermore, since the generated counterfactual labels correspond largely to the minority classes, the relatively high minority class $F_1$ scores suggest that the generated counterfactuals are sufficiently realistic to allow the model to learn.

Taken together, these results suggest that \textit{counterfactual augmentation is indeed successful in correcting presentation bias---and that it does a better job than existing bias-correction methods}. From an empirical standpoint, counterfactual augmentation is both useful and stable across settings and tasks, yielding consistently good performance.



\subsection{Model analysis: Why does counterfactual augmentation work?}
\label{sec:model_analysis}

Counterfactual augmentation produces clear empirical gains in downstream performance over both uncorrected models and existing bias-correction methods. In this section, we analyze our generated counterfactuals to better understand these improvements. Although true counterfactuals are never known in the real world, in these experiments we do have access to the true counterfactuals of $D_{train}$, which we withheld in the process of inducing presentation bias. We use these  as a basis for comparison with our generated counterfactuals.

\begin{figure*}[!ht]
    \centering
    \scalebox{1.0}[0.93]{\includegraphics[width=0.28\textwidth]{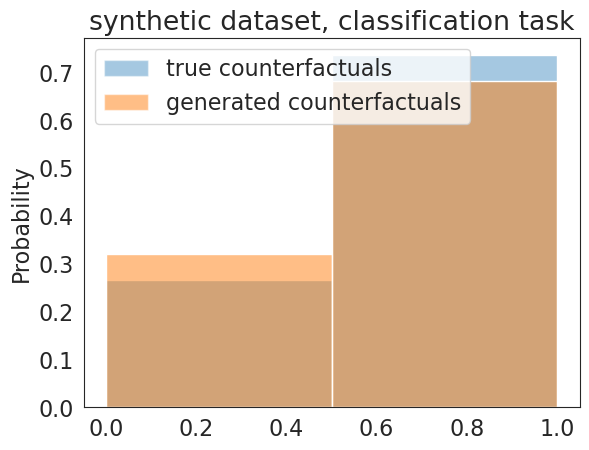}}
    \scalebox{1.0}[0.93]{\includegraphics[width=0.28\textwidth]{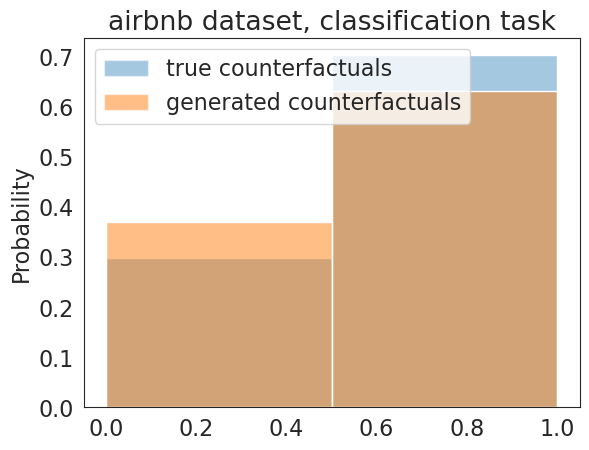}}
    \scalebox{1.0}[0.93]{\includegraphics[width=0.28\textwidth]{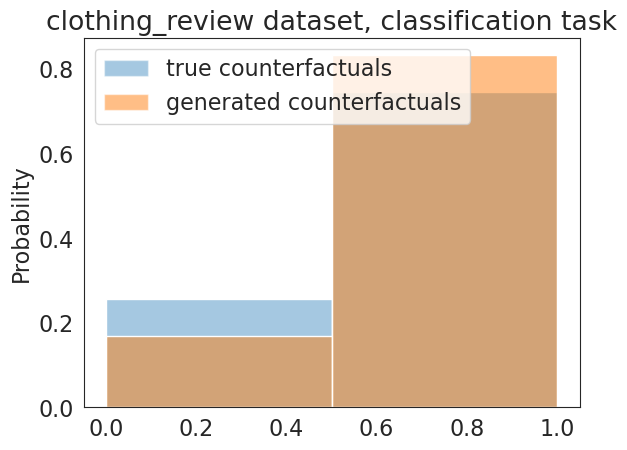}} \\
    \scalebox{1.0}[0.93]{\includegraphics[width=0.28\textwidth]{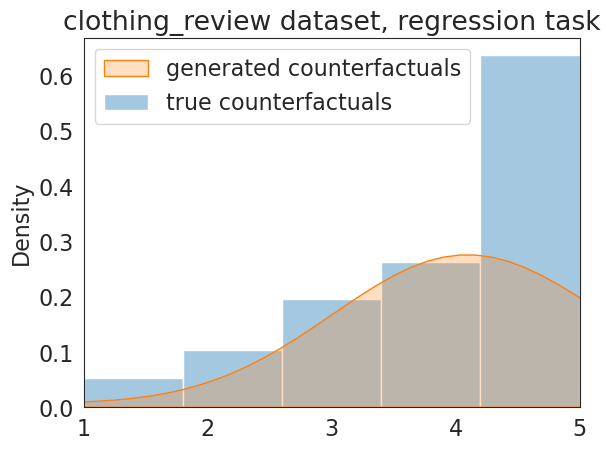}}
    \scalebox{1.0}[0.93]{\includegraphics[width=0.28\textwidth]{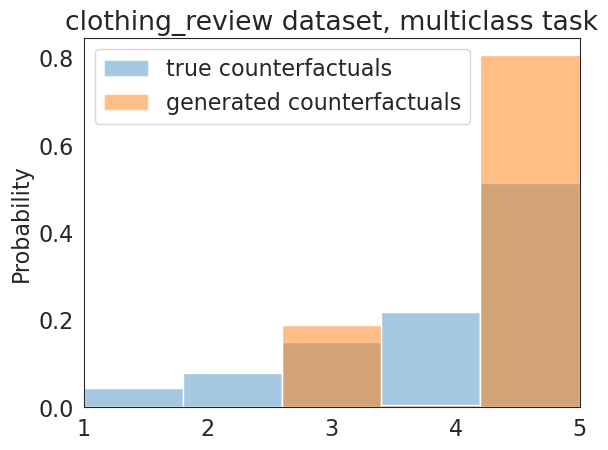}}
    \scalebox{1.0}[0.93]{\includegraphics[width=0.28\textwidth]{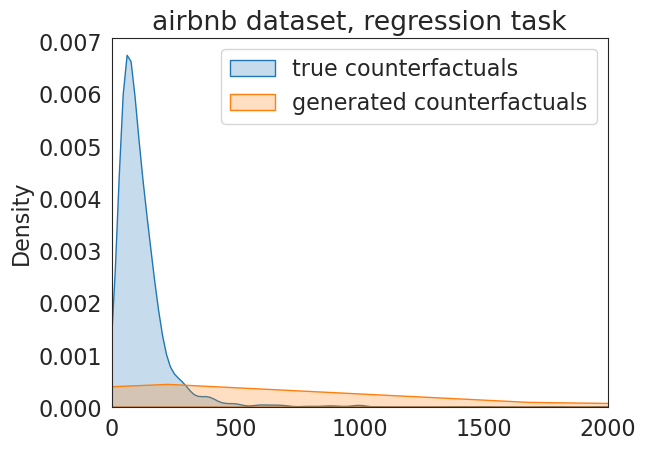}}
    \caption{Comparison of the distributions of the generated counterfactuals and the true counterfactuals.}
    \label{fig:synthetic_real_label_dist}
\end{figure*}

\begin{figure*}[!ht]
    \centering
    \scalebox{1.0}[0.93]{\includegraphics[width=0.28\textwidth]{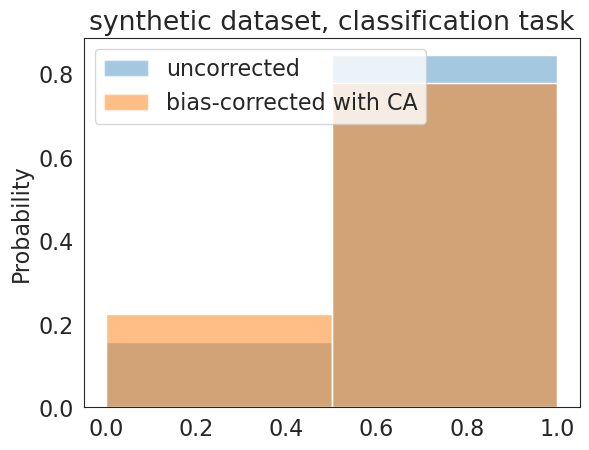}}
    \scalebox{1.0}[0.93]{\includegraphics[width=0.28\textwidth]{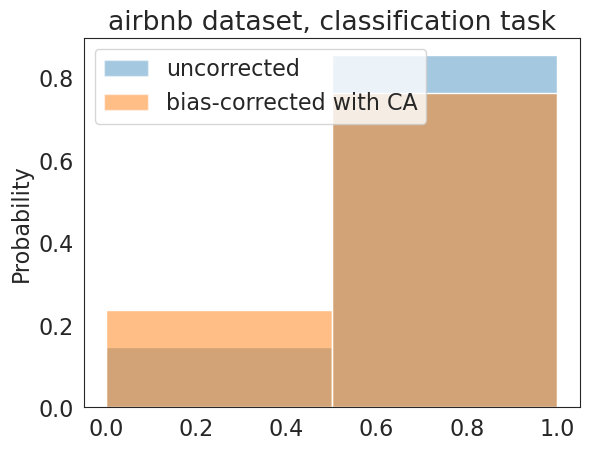}}
    \scalebox{1.0}[0.93]{\includegraphics[width=0.28\textwidth]{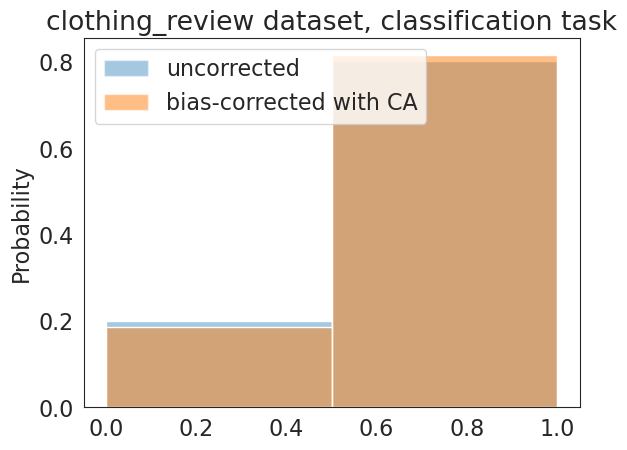}} \\
    \scalebox{1.0}[0.93]{\includegraphics[width=0.28\textwidth]{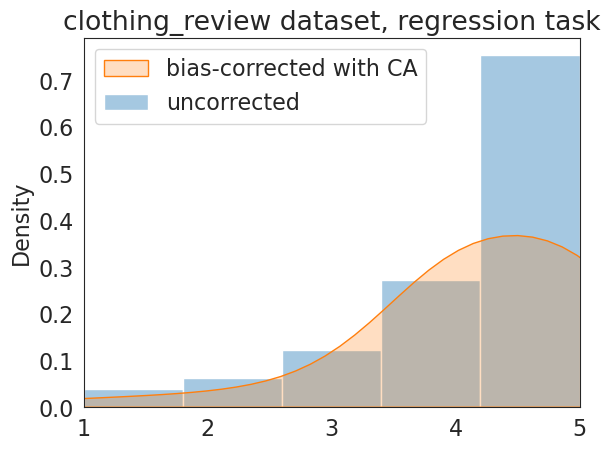}}
    \scalebox{1.0}[0.93]{\includegraphics[width=0.28\textwidth]{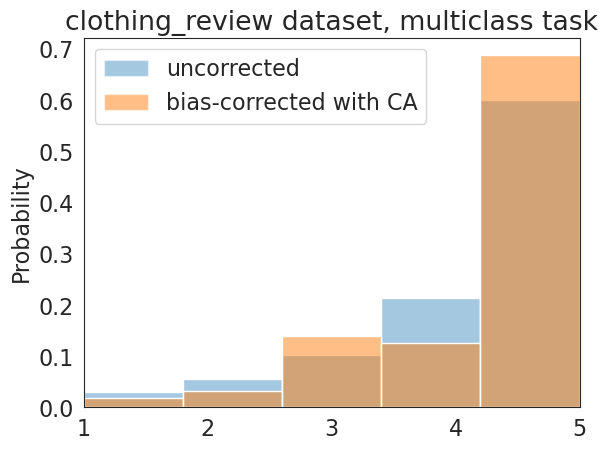}}
    \scalebox{1.0}[0.93]{\includegraphics[width=0.28\textwidth]{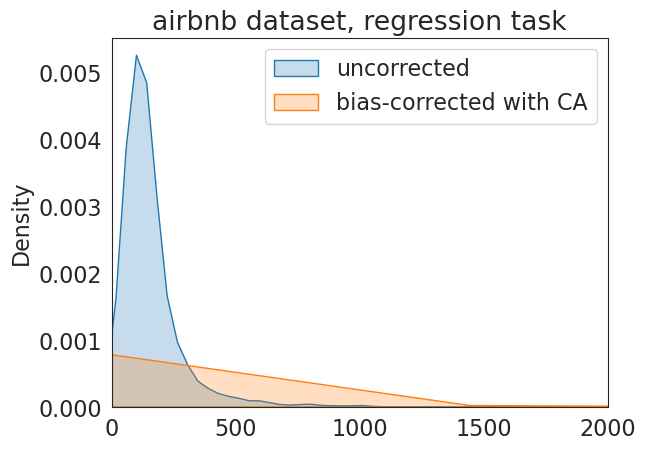}}
    \caption{Comparison of uncorrected label distributions and label distributions after bias-correction with CA.}
    \label{fig:label_balance}
\end{figure*}

\textbf{Comparing counterfactual distributions.} To assess how well our generated counterfactuals correspond to real counterfactuals, we plot their distributions together for each combination of dataset and task (Figure \ref{fig:synthetic_real_label_dist}). We observe that for the easier binary classification task, \textit{the distribution of generated counterfactuals closely reflects that of the true counterfactuals across all data settings}. On these tasks, it appears that the generated counterfactuals are a good approximation of the true counterfactuals. For the more difficult multi-class classification and regression tasks, the difference between the generated and true distributions is greater. 

Reducing presentation bias helps correct the label imbalance that exists in the overall label distribution (Figure \ref{fig:label_balance}). However, the generated counterfactual distribution also tends to be more uniform compared to the true counterfactual distribution (seen in 4 out of 6 plots in Figure \ref{fig:synthetic_real_label_dist}). Therefore, even aside from the reduced presentation bias, the greater uniformity of the generated counterfactuals may further correct label imbalance. In general, we observe that \textit{the bias-corrected label distribution $P_{CA}(Y)$ is more balanced than the uncorrected label distribution $P(Y|A=1)$.} This reduction in label imbalance better enables a model to learn on the bias-corrected set.



\textbf{Performance of an oracle.} Because we have access to the true counterfactuals, we can train an oracle model over an unbiased version of $D_{train}$. By comparing the oracle to counterfactual augmentation, we can determine how well counterfactual augmentation recovers performance compared to the original unbiased data. We report the results of the oracle in Tables \ref{tab:results_classification_unbiased} through \ref{tab:results_regression_multiclass_biased}.

For most tasks and data settings, we observe that---as expected---counterfactual augmentation still results in some loss of performance compared to the oracle.\footnote{In some cases, we see that the oracle achieves worse performance than CA. We posit that the tendency of the generated counterfactual distribution toward the uniform ``over-corrects'' label imbalance, so the bias-corrected label distribution is more uniform than the unbiased label distribution. This may make model training easier on the bias-corrected distribution (e.g., on sparser portions of the unbiased distribution).} However, the performance gap between CA and the oracle is generally substantially less than the performance gap between CA and the next-best bias-correction method. These results suggest that although CA constitutes a significant improvement over existing methods, further refinement of the counterfactual generation method may be able to yield even better results.



\section{Related Work}
\label{sec:related_work}

Presentation bias may be considered a type of \textit{selection bias}, in which the sampling distribution differs from the population distribution. Selection bias is a core challenge of observational causal inference, where the causal effect of a treatment $A$ on an outcome $Y$ is estimated not from a randomized trial but from observed data. Since the treatment assignment mechanism is not random, it must be accounted for during estimation.


One common method for addressing selection bias in causal inference is \textit{inverse propensity weighting} (IPW) \citep{robins1994estimation, hernan2006estimating}. At a high level, IPW up-weights samples corresponding to treatment conditions that are unlikely to be observed and down-weights samples corresponding to treatment conditions that are likely to be observed, such that all treatment conditions appear to be equally likely over the data distribution. This blocks the causal path between the treatment assignment and the outcome. 



\textbf{IPW for presentation bias correction.} Using this principle, a number of works propose an inverse propensity weighted empirical loss function that can be used to reduce the effects of presentation bias when training a model on biased data \citep{wang2016learning, schnabel2016recommendations, joachims2017unbiased}. Several works also engage with IPW in more complex ways. \citet{krauth2022breaking} address a longitudinal bias setting and propose an algorithm that maximizes the desired outcome at each time step using an IPW-based estimator. \citet{shi2019adapting} introduce Dragonnet, a fully-connected multi-head neural network that jointly predicts the treatment and the outcome, simultaneously yielding both a propensity score estimate and a predicted outcome.

\textbf{Task-based presentation bias correction.} Because presentation bias can appear in many task settings, there exist a number of task-specific approaches for reducing presentation bias. In information retrieval, for example, unbiased learners of click data \citep{ai2018unbiased} and propensity-weighted rank metrics \citep{agarwal2019general} have been proposed, while in the recommender literature, methods have been developed for the matrix factorization setting \citep{bonner2018causal, wang2020causal}. However, the task-specific nature of these methods limits their generalizability compared to counterfactual augmentation.

\textbf{Estimating counterfactuals.} The inability to know an individual's counterfactual is a central challenge of causal inference. However, recent works in the deep learning literature have made large inroads toward estimating individual treatment effects \citep{shalit2017estimating, louizos2017causal, yoon2018ganite}, which is an adjacent task to estimating individual counterfactuals. We draw upon this body of work as a basis for obtaining high-quality counterfactuals.

\textbf{Counterfactuals in NLP.} Our work is contextualized within a recent body of research that has shown that counterfactuals are an effective supplement to training data when learning language models \cite{wang2021spurious, qian2021counterfactual, yang2021exploring, howard2022neurocounterfactuals}. Existing works largely rely on manually created counterfactuals or programmatically generated counterfactuals. Our method advances beyond prior works by leveraging the causal mechanism behind the missing portions of the data distribution to efficiently generate targeted, high-quality counterfactuals.


\section{Conclusion}
\vspace{-2mm}
\label{sec:conclusion}
In this paper, we introduced \textit{counterfactual augmentation}, a causal method for correcting presentation bias using generated counterfactuals. We described the causal mechanism behind presentation bias in real-world machine learning systems that rely on user feedback, and we explained the causal reasoning behind counterfactual augmentation. We presented empirical evaluations using counterfactual augmentation to reduce presentation bias, and we found that our approach significantly outperforms existing methods. Finally, we conducted a model analysis to explore why counterfactual augmentation is effective in addressing presentation bias. Given the prevalence of presentation bias in real-world deployments of machine learning models, our findings suggest that counterfactual augmentation has the potential to improve the quality of user-facing machine learning models across many types of applications.

\section{Acknowledgements}

This material is based upon work partially supported by Microsoft, the National Science Foundation (awards 1722822 and 1750439), and the National Institutes of Health (awards R01MH125740, R01MH132225, R01MH096951, and R21MH130767). Victoria Lin is partially supported by a Meta Research PhD Fellowship. Any opinions, findings, conclusions, or recommendations expressed in this material are those of the author(s) and do not necessarily reflect the views of the sponsors, and no official endorsement should be inferred. We are grateful to Alan Thomas and Sebastian de la Chica for many helpful discussions and feedback.

\section{Limitations}

The most significant limitation of counterfactual augmentation is its requirement that the generated counterfactuals be sufficiently close to true counterfactuals; otherwise, the counterfactually augmented data distribution $P_{CA}(X,Y)$ is not a good approximation of the true data distribution $P(X,Y)$. If poor-quality counterfactuals are produced and $P_{CA}(X,Y)$ is very different from $P(X,Y)$, counterfactual augmentation could instead hurt models that are trained on the augmented data. Although our multimodal counterfactual GAN generates high-quality counterfactuals for the tasks and data settings that we evaluate, we do not know if this will be the case across every task and data setting. A different counterfactual estimation method may be required depending on the particular problem.

Based on failure modes of causal effect estimation in statistical causal inference, we hypothesize that lower-quality counterfactuals may be produced if:
\begin{itemize}
    \item The causal mechanism of presentation bias is misspecified.
    \item The feature data is very noisy or sparse, making it difficult to learn counterfactuals.
    \item The counterfactual generation model does not have enough capacity to model the data (could be more of a problem for “traditional” statistical linear models).
\end{itemize}

\section{Ethics Statement}

\textbf{Broader impact.} Deep learning models have been shown to perpetuate and even amplify the biases in their training data \citep{bolukbasi2016man, swinger2019biases, caliskan2017semantics}. Often, these biases manifest in a similar way to presentation bias: that is, only a portion of the theoretical data distribution is contained in the model's training dataset, which impacts what the model learns.

Therefore, we believe that counterfactual augmentation may be helpful not only in correcting presentation bias but also in reducing social biases in data. In principle, counterfactual augmentation can be used to correct any type of bias for which the causal mechanism is known. The causal mechanism is used to generate counterfactuals, which augment the unobserved portion of the data distribution. Consequently, counterfactual augmentation may also be helpful in correcting social biases and helping make data more fair.

\textbf{Ethical considerations.} When used in conjunction with multimodal data, as it is in this paper, counterfactual augmentation relies in part on large pre-trained models to generate counterfactuals. As a result, it is also possible that the generated counterfactuals themselves may encode the biases contained in large pre-trained models. Users should be cautious when employing counterfactual augmentation in sensitive settings or when using it to reduce biases on protected attributes. 

Additionally, we acknowledge the environmental impact of large language and image models, which are used in this work.


\bibliography{ref}
\bibliographystyle{acl_natbib}

\appendix


\section{Experimental Details}
\label{sec:appendix}

\subsection{Data}
\label{sec:data_details}

Details of our data splits are provided in Table \ref{tab:data_details}, including dataset composition and licensing information. Both Airbnb and Clothing are publicly available datasets, and the Synthetic dataset is available on our GitHub. This release of the Synthetic dataset---for research purposes only---is compatible with the intended use of the Airbnb dataset, on which it is based. All datasets are in English.

\begin{table*}[!ht]
    \centering
    \begin{tabular}{c|ccccc|c}
    \toprule
    \toprule
    Dataset & $D_{original}$ & $D_{train}$ & $D_{eval}$ & $D_{biased}$ & n$_{total}$ & License \\
    \midrule
    Synthetic & 4,572 & 5,400 & 9,135 & 5,400 & 22,895 & CC0 1.0 \\
    Airbnb & 4,572 & 5,400 & 9,135 & 5,400 & 22,895 & CC0 1.0 \\
    Clothing & 4,698 & 5,400 & 9,394 & 5,400 & 23,486 & CC0 1.0 \\
    \bottomrule
    \bottomrule
    \end{tabular}
    \caption{Composition of data splits. For each dataset, the number of samples in $D_{original}$, $D_{train}$, $D_{eval}$, and $D_{biased}$ is given, along with total samples for each dataset. Licensing information is also provided.}
    \label{tab:data_details}
\end{table*}

\subsection{Training details}
\label{sec:training_details}

Our use of all pre-existing models is limited to research purposes only and is compliant with their intended use. 

\textbf{Language and image models.} Our language and image transformer models were built on the HuggingFace\footnote{\url{https://huggingface.co/}} \verb|transformers| library (version 4.18.0), with pre-trained models taken from the HuggingFace model hub. For fine-tuning, we used an Adam optimizer and learning rates [$2 \times 10^{-1}, 2 \times 10^{-2}, 2 \times 10^{-3}, 2 \times 10^{-4}, 2 \times 10^{-5}$], and we found $2 \times 10^{-5}$ to be the best learning rate across all models. We trained for 5 epochs and selected the model with the best validation loss. All other hyperparameters were set to Trainer class defaults from the \verb|transformers| library.

\textbf{Dragonnet.} Our implementation of Dragonnet was based on its public code release,\footnote{\url{https://github.com/claudiashi57/dragonnet}} which uses Tensorflow (version 2.8.0). All hyperparameters were kept to their default values from the original code.

\textbf{Multimodal counterfactual GAN.} Our implementation of our multimodal counterfactual GAN uses PyTorch (version 1.11.0) and was built on skeleton code from the public code release\footnote{\url{https://github.com/vanderschaarlab/mlforhealthlabpub/tree/main/alg/ganite}} of GANITE. We tune over the following hyperparameters (best in bold):
\begin{itemize}[noitemsep]
    \item Hidden layer size of generator and discriminator: [\textbf{128}, 256, 512, 1024]
    \item Number of generator iterations: [200, \textbf{500}, 1000, 2000]
    \item Number of discriminator iterations: [5, 6, 7, \textbf{8}, 9, 10]
    \item Learning rate: [$10^{-1}$, $10^{-2}$, $10^{-3}$, $10^{-4}$, $\mathbf{10^{-5}}$]
    \item Separate discriminators: [\textbf{yes}, no]
    \item Scaling data: [yes, \textbf{no}\footnote{For the clothing multi-class classification task only, scaling was the better option.}]
    \item Fine-tune encoders and GAN sequentially: [\textbf{yes}, no]
\end{itemize}

The number of parameters and licenses for each of the models is reported in Table \ref{tab:num_params}.

\begin{table}[!ht]
    \centering
    \resizebox{\columnwidth}{!}{
    \begin{tabular}{c|c|c}
    \toprule
    \toprule
        Model & \# parameters & License \\
    \midrule
        Dragonnet & 474,603 & Unknown \\
        DistilBERT & 66,955,010 & Apache-2.0 \\
        ViT & 85,800,194 & Apache-2.0 \\
        Counterfactual GAN & 500,601 & BSD-3-Clause \\
    \bottomrule
    \bottomrule
    \end{tabular}
    }
    \caption{Number of parameters and license for each model. Note that because the best-performing configuration for the counterfactual GAN trains the language and image encoders and the GAN sequentially, the number of parameters for the counterfactual GAN excludes the encoder parameters. When the encoders are fine-tuned in parallel with the GAN, we consider the number of parameters for the counterfactual GAN to be 153,713,529.}
    \label{tab:num_params}
\end{table}

\subsection{Runtimes}

\begin{table}[!ht]
    \centering
    \begin{tabular}{c|cc}
    \toprule
    \toprule
        & \multicolumn{2}{c}{Synthetic} \\
        & Runtime & Added time \\
        \midrule
        Uncorrected & 1304 & - \\
        IPW & 1350 & 46 \\
        Dragonnet & 1494 & 190 \\
        CA (ours) & 1361 & 57 \\
    \bottomrule
    \bottomrule
    \end{tabular}
    \caption{Runtimes (in seconds) of all models and bias-correction methods for the Synthetic binary classification task. ``Added time'' denotes the additional time the bias-correction method requires relative to the uncorrected model.}
    \label{tab:runtimes_synthetic}
\end{table}

\begin{table*}[!ht]
    \centering
    \begin{tabular}{c|cc|cc}
    \toprule
    \toprule
        & \multicolumn{4}{c}{Airbnb} \\
        \cmidrule{2-5}
        & \multicolumn{2}{c|}{Binary classification} & \multicolumn{2}{c}{Regression} \\
        & Runtime & Added time & Runtime & Added time \\
        \midrule
        Uncorrected & 1319 & - & 1320 & - \\
        IPW & 1367 & 48 & 1431 & 111 \\
        Dragonnet & 1423 & 104 & - & - \\
        CA (ours) & 1383 & 64 & 1334 & 14 \\
    \bottomrule
    \bottomrule
    \end{tabular}
    \caption{Runtimes (in seconds) of all models and bias-correction methods for the Airbnb tasks. ``Added time'' denotes the additional time the bias-correction method requires relative to the uncorrected model.}
    \label{tab:runtimes_airbnb}
\end{table*}

\begin{table*}[!ht]
    \centering
    \begin{tabular}{c|cc|cc|cc}
    \toprule
    \toprule
        & \multicolumn{6}{c}{Clothing} \\
        \cmidrule{2-7}
        & \multicolumn{2}{c|}{Binary classification} & \multicolumn{2}{c|}{Multi-class classification} & \multicolumn{2}{c}{Regression} \\
        & Runtime & Added time & Runtime & Added time & Runtime & Added time \\
        \midrule
        Uncorrected & 251 & - & 246 & - & 247 & - \\
        IPW & 272 & 21 & 271 & 25 & 273 & 26 \\
        Dragonnet & 351 & 100 & - & - & - & - \\
        CA (ours) & 265 & 15 & 260 & 14 & 317 & 71 \\
    \bottomrule
    \bottomrule
    \end{tabular}
    \caption{Runtimes (in seconds) of all models and bias-correction methods for the Airbnb tasks. ``Added time'' denotes the additional time the bias-correction method requires relative to the uncorrected model.}
    \label{tab:runtimes_clothingreview}
\end{table*}

Counterfactual augmentation is computationally efficient, and its computational overhead is minimal compared to the computation required to learn task models. We provide runtimes of all models and bias-correction methods for each of our evaluation tasks in Tables \ref{tab:runtimes_synthetic}, \ref{tab:runtimes_airbnb}, and \ref{tab:runtimes_clothingreview}. For every task, either IPW or counterfactual augmentation is the most efficient bias-correction method.

\subsection{Computing resources}

A portion of our experiments were conducted using machines with consumer-level NVIDIA graphics cards. Our remaining experiments were conducted using cloud computing resources. We estimate the number of GPU hours used to be around 150.

\section{Additional Experiments}

\subsection{Counterfactual augmentation without feature evolution}
\label{sec:tabonly_results}

\begin{table*}[!ht]
    \centering
    \begin{tabular}{c|cccc|cccc|cccc}
    \toprule
    \toprule
        & \multicolumn{4}{c|}{Synthetic} & \multicolumn{4}{c|}{Airbnb} & \multicolumn{4}{c}{Clothing} \\
        & Acc. & $F_1$ & ${F_1}_{mac}$ & ${F_1}_{min}$ & Acc. & $F_1$ & ${F_1}_{mac}$ & ${F_1}_{min}$ & Acc. & $F_1$ & ${F_1}_{mac}$ & ${F_1}_{min}$ \\
        \midrule
        Uncorrected & 79.7 & 70.8 & 44.6 & 0.5 & 86.7 & 86.1 & 77.2 & 62.5 & 77.1 & 67.2 & 43.8 & 0.5 \\
        IPW & 82.0 & 78.6 & 62.7 & 35.8 & \textbf{86.7} & 86.5 & 78.4 & 64.9 & 77.1 & 67.2 & 43.8 & 0.5 \\
        Dragonnet & 82.2 & 79.3 & 63.5 & 37.5 & 81.6 & 79.2 & 64.6 & 40.1 & 80.1 & 76.5 & 63.0 & 37.8 \\
        CA (ours) & \textbf{82.7} & \textbf{81.5} & \textbf{69.7} & \textbf{49.9} & 86.2 & \textbf{86.6} & \textbf{79.4} & \textbf{67.5} & \textbf{80.3} & \textbf{76.7} & \textbf{63.0} & \textbf{37.9} \\
        \midrule
        Improvement & 0.5\% & 2.2\% & 6.2\% & 12.4\% & - & 0.1\% & 1.0\% & 2.6\% & 0.2\% & 0.2\% & 0.0\% & 0.1\% \\
    \bottomrule
    \bottomrule
    \end{tabular}
    \caption{Results on binary classification tasks without feature evolution (unbiased evaluation dataset).}
    \label{tab:tabonly_results_classification_unbiased}
\end{table*}

\begin{table*}[!ht]
    \centering
    \begin{tabular}{c|cccc|cccc|cccc}
    \toprule
    \toprule
        & \multicolumn{4}{c|}{Synthetic} & \multicolumn{4}{c|}{Airbnb} & \multicolumn{4}{c}{Clothing} \\
        & Acc. & $F_1$ & ${F_1}_{mac}$ & ${F_1}_{min}$ & Acc. & $F_1$ & ${F_1}_{mac}$ & ${F_1}_{min}$ & Acc. & $F_1$ & ${F_1}_{mac}$ & ${F_1}_{min}$ \\
        \midrule
        Uncorrected & 84.7 & 77.7 & 46.2 & 0.7 & \textbf{89.1} & \textbf{88.4} & 74.6 & 55.5 & 79.6 & 70.7 & 44.5 & 0.4 \\
        IPW & 85.0 & 80.0 & 53.5 & 15.2 & 88.6 & 88.4 & 75.2 & 56.9 & 79.6 & 70.7 & 44.5 & 0.4 \\
        Dragonnet & 84.9 & 78.9 & 49.8 & 7.8 & 85.7 & 80.9 & 51.9 & 11.6 & \textbf{80.7} & \textbf{74.9} & \textbf{54.6} & \textbf{20.2} \\
        CA (ours) & \textbf{85.0} & \textbf{81.7} & \textbf{59.2} & \textbf{26.7} & 88.3 & 88.2 & \textbf{75.2} & \textbf{57.2} & 80.5 & 74.6 & 53.6 & 18.2 \\
    \midrule
        Improvement & - & 1.7\% & 5.7\% & 11.5\% & - & - & 0.0\% & 0.3\% & - & - & - & - \\
    \bottomrule
    \bottomrule
    \end{tabular}
    \caption{Results on binary classification tasks without feature evolution (biased evaluation dataset).}
    \label{tab:tabonly_results_classification_biased}
\end{table*}

\begin{table*}[!ht]
    \centering
    \begin{tabular}{c|cc|cc|ccc}
    \toprule
    \toprule
        & \multicolumn{4}{c|}{Regression} & \multicolumn{3}{c}{Multi-class} \\
        \cmidrule{2-8}
        & \multicolumn{2}{c|}{Airbnb} & \multicolumn{2}{c|}{Clothing} & \multicolumn{3}{c}{Clothing} \\
        & $R^2$ & RMSE & $R^2$ & RMSE & Acc. & $F_1$ & ${F_1}_{mac}$ \\
        \midrule
        Uncorrected & 0.079 & 0.959 & -0.419 & 1.191 & 55.6 & 39.8 & 14.3 \\
        IPW & 0.083 & 0.958 & -0.419 & 1.191 & 55.6 & 39.8 & 14.3 \\
        Dragonnet & - & - & - & - & - & - & - \\
        CA (ours) & \textbf{0.269} & \textbf{0.855} & \textbf{-0.133} & \textbf{1.064} & \textbf{57.2} & \textbf{44.6} & \textbf{20.2} \\
        \midrule
        Improvement & 18.6\% & 10.3\% & 28.6\% & 12.7\% & 1.6\% & 4.8\% & 5.9\% \\
    \bottomrule
    \bottomrule
    \end{tabular}
    \caption{Results on regression and multi-class classification tasks without feature evolution (unbiased evaluation dataset). Our RMSE metric is normalized RMSE, or RMSE divided by the standard deviation of the evaluation set.}
    \label{tab:tabonly_results_regression_multiclass_unbiased}
\end{table*}

\begin{table*}[!ht]
    \centering
    \begin{tabular}{c|cc|cc|ccc}
    \toprule
    \toprule
        & \multicolumn{4}{c|}{Regression} & \multicolumn{3}{c}{Multi-class} \\
        \cmidrule{2-8}
        & \multicolumn{2}{c|}{Airbnb} & \multicolumn{2}{c|}{Clothing} & \multicolumn{3}{c}{Clothing} \\
        & $R^2$ & RMSE & $R^2$ & RMSE & Acc. & $F_1$ & ${F_1}_{mac}$ \\
        \midrule
        Uncorrected & 0.041 & 0.979 & -0.329 & 1.153 & 58.4 & 43.0 & 14.7 \\
        IPW & 0.043 & 0.979 & -0.329 & 1.153 & 58.4 & 43.0 & 14.7 \\
        Dragonnet & - & - & - & - & - & - & - \\
        CA (ours) & \textbf{0.207} & \textbf{0.891} & \textbf{-0.130} & \textbf{1.063} & \textbf{58.6} & \textbf{43.9} & \textbf{16.1} \\
        \midrule
        Improvement & 16.4\% & 8.8\% & 19.9\% & 9.0\% & 0.2\% & 0.9\% & 1.4\% \\
    \bottomrule
    \bottomrule
    \end{tabular}
    \caption{Results on regression and multi-class classification tasks without feature evolution (biased evaluation dataset).}
    \label{tab:tabonly_results_regression_multiclass_biased}
\end{table*}

In this section, we present results (Tables \ref{tab:tabonly_results_classification_unbiased}, \ref{tab:tabonly_results_classification_biased}, \ref{tab:tabonly_results_regression_multiclass_unbiased}, and \ref{tab:tabonly_results_regression_multiclass_biased}) that verify that counterfactual augmentation does not require feature evolution to successfully correct presentation bias. We follow the same experimental procedures as in Section \ref{sec:experiments}; however, for all three datasets---Synthetic, Airbnb, and Clothing---we use only tabular features for all data splits $D_{original}$, $D_{train}$, $D_{eval}$, and $D_{biased}$.

We observe that as is the case in our main results, counterfactual augmentation generally produces improvements in overall performance and macro/minority class performance, both relative to the uncorrected baseline and to the competing bias-correction baselines, IPW and Dragonnet.

Performance of both the baselines and counterfactual augmentation is less consistent compared to the feature evolution setting (this is particularly evident in the clothing regression task, where $R^2$ is negative). We hypothesize that there is not sufficient information encoded in the tabular data to learn certain tasks well. Furthermore, it is also more difficult to generate the counterfactuals for counterfactual augmentation, since the GAN now also has access only to tabular data.

\begin{subsection}{Modality ablations}

\begin{table*}[!ht]
    \centering
    \begin{tabular}{c|cccc|cccc}
    \toprule
    \toprule
        & \multicolumn{4}{c|}{Synthetic} & \multicolumn{4}{c}{Airbnb}\\
        & Acc. & $F_1$ & ${F_1}_{mac}$ & ${F_1}_{min}$ & Acc. & $F_1$ & ${F_1}_{mac}$ & ${F_1}_{min}$ \\
        \midrule
        Uncorrected & 79.7 & 70.8 & 44.5 & 0.4 & 82.9 & 78.8 & 61.4 & 32.6  \\
        IPW & 81.8 & 77.5 & 59.8 & 30.1 & 86.3 & 85.7 & 76.5 & 61.3  \\
        Dragonnet & 80.6 & 74.8 & 53.2 & 17.3 & 81.5 & 78.8 & 63.4 & 37.6 \\
        CA (ours) & \textbf{82.2} & \textbf{81.0} & \textbf{68.9} & \textbf{48.6} & \textbf{87.4} & \textbf{87.7} & \textbf{81.2} & \textbf{70.3}  \\
        \midrule
        Improvement & 0.4\% & 3.5\% & 9.1\% & 18.4\% & 1.1\% & 2.1\% & 4.7\% & 9.0\% \\
    \bottomrule
    \bottomrule
    \end{tabular}
    \caption{Results on binary classification tasks with the language modality only (unbiased evaluation dataset).}
    \label{tab:textonly_results_classification_unbiased}
\end{table*}

\begin{table*}[!ht]
    \centering
    \begin{tabular}{c|cccc|cccc}
    \toprule
    \toprule
        & \multicolumn{4}{c|}{Synthetic} & \multicolumn{4}{c}{Airbnb}\\
        & Acc. & $F_1$ & ${F_1}_{mac}$ & ${F_1}_{min}$ & Acc. & $F_1$ & ${F_1}_{mac}$ & ${F_1}_{min}$ \\
        \midrule
        Uncorrected & 84.6 & 77.6 & 45.8 & 0.0 & 87.4 & 83.8 & 59.9 & 26.7  \\
        IPW & \textbf{84.9} & 79.3 & 51.0 & 10.2 & 88.3 & 87.5 & 72.2 & 51.1  \\
        Dragonnet & 84.7 & 78.0 & 46.8 & 1.9 & 86.3 & 80.9 & 51.2 & 9.8 \\
        CA (ours) & 84.6 & \textbf{81.3} & \textbf{58.4} & \textbf{25.4} & \textbf{88.6} & \textbf{88.6} & \textbf{76.4} & \textbf{59.5}  \\
        \midrule
        Improvement & - & 2.0\% & 7.4\% & 15.2\% & 0.3\% & 1.2\% & 4.2\% & 8.5\% \\
    \bottomrule
    \bottomrule
    \end{tabular}
    \caption{Results on binary classification tasks with the language modality only (biased evaluation dataset).}
    \label{tab:textonly_results_classification_biased}
\end{table*}

\begin{table*}[!ht]
    \centering
    \begin{tabular}{c|cc|cc}
    \toprule
    \toprule
        & \multicolumn{2}{c|}{Airbnb (unbiased)} & \multicolumn{2}{c}{Airbnb (biased)} \\
        & $R^2$ & RMSE & $R^2$ & RMSE  \\
        \midrule
        Uncorrected & 0.031 & 0.984 & -0.007 & 1.004 \\
        IPW & 0.038 & 0.981 & -0.005 & 1.002 \\
        Dragonnet & - & - & - & - \\
        CA (ours) & \textbf{0.224} & \textbf{0.881} & \textbf{0.200} & \textbf{0.894} \\
        \midrule
        Improvement & 18.5\% & 10.0\% & 20.5\% & 10.8\% \\
    \bottomrule
    \bottomrule
    \end{tabular}
    \caption{Results on regression and multi-class classification tasks with the language modality only (unbiased evaluation dataset). Our RMSE metric is normalized RMSE, or RMSE divided by the standard deviation of the evaluation set.}
    \label{tab:textonly_results_regression}
\end{table*}

To further demonstrate the utility of counterfactual augmentation in natural language settings, in this section we report results from experiments with ablated (language-only) versions of the Airbnb and Synthetic datasets (Tables \ref{tab:textonly_results_classification_unbiased}, \ref{tab:textonly_results_classification_biased}, \ref{tab:textonly_results_regression}). Again, we follow the same experimental procedures as in Section \ref{sec:experiments}, but we make only language features available in $D_{train}$, $D_{eval}$, and $D_{biased}$.

We find that---consistent with our other results---counterfactual augmentation continues to outperform uncorrected models and existing bias-correction methods. Interestingly, we observe that bias correction appears to work better in some data settings given only the text modality, relative to models that have access to both text and images.

\end{subsection}

\end{document}